\pdfoutput=1

\documentclass[11pt]{article}

\usepackage[preprint]{acl}

\usepackage{times}
\usepackage{latexsym}

\usepackage[T1]{fontenc}

\usepackage[utf8]{inputenc}

\usepackage{microtype}

\usepackage{inconsolata}

\usepackage{graphicx}

\usepackage{soul}

\title{Local Transcription Models in Home Care Nursing in Switzerland: an Interdisciplinary Case Study}

\author{
  \textbf{Jeremy Kramer\textsuperscript{1}},
  \textbf{Tetiana Kravchenko\textsuperscript{2}},
  \textbf{Beatrice Kaufmann\textsuperscript{3}},
  \textbf{Friederike J.S. Thilo\textsuperscript{1}},
\\
 \textbf{Mascha Kurpicz-Briki\textsuperscript{2}},
\\
\\
  \textsuperscript{1}Applied Research \& Development in Nursing, Bern University of Applied Sciences, Bern, Switzerland \\
  \textsuperscript{2}Generative AI Lab, Bern University of Applied Sciences, Biel, Switzerland \\
  \textsuperscript{3}Institute of Design Research, Bern Academy of the Arts, Bern, Switzerland \\
  \small{
    \textbf{Correspondence:} \href{mailto:mascha.kurpicz@bfh.ch}{mascha.kurpicz@bfh.ch}
  }
}

\begin{document}
\maketitle
\begin{abstract}
Latest advances in the field of natural language processing (NLP) enable new use cases for different domains, including the medical sector. In particular, transcription can be used to support automation in the nursing documentation process and give nurses more time to interact with the patients. However, different challenges including (a) data privacy, (b) local languages and dialects, and (c) domain-specific vocabulary need to be addressed. In this case study, we investigate the case of home care nursing documentation in Switzerland. We assessed different transcription tools and models, and conducted several experiments with OpenAI Whisper, involving different variations of German (i.e., dialects, foreign accent) and manually curated example texts by a domain expert of home care nursing. Our results indicate that even the used out-of-the-box model performs sufficiently well to be a good starting point for future research in the field. 
\end{abstract}

\section{Introduction}

The topic of transcription in healthcare documentation has gained increasing relevance, particularly with the recent advancements in transformer-based models, which have opened up new possibilities for downstream natural language processing (NLP) applications. However, there can be challenges in the practical implementation of such models in the healthcare systems, due to different technical and regulatory restrictions. From the results of different recent systematic literature surveys and studies on applications for speech recognition and simultaneous recording of patient-nursing and patient-physician interactions \cite{falcetta2023automatic} \cite{dinari2023benefits} \cite{kumar2024comprehensive}, the following list of challenges can be derived: (a) the complexity of (domain-specific) vocabulary and relevant parts identification, and contextual information, (b) legal, ethical and moral challenges, (c) transcription of dialects or accents, or variability in speaking styles, (d) environmental factors (e.g., noisy background, data integration with electronic health records, real-time processing), and (e) practical implementation (e.g., education and training, need for manual correction). Transcription can be used for different tasks and is dependent on the local context with regard to these challenges.

In healthcare applications, the local languages spoken by the patients, nurses and physicians are relevant for the technical implementation. For example, medical drug prescriptions acquired on smartphones through spoken dialogue in the French language have been investigated \cite{kocabiyikoglu-etal-2022-spoken}. The authors state that in their domain there is a lack of speech corpora, as most available resources are in text form and in English. This is a general limitation of NLP research, lacking diversity in terms of languages \cite{joshi-etal-2020-state} and thus not sufficiently investigating the challenges of languages other than English, and making corresponding resources available. This is in particular also the case for transcription in nursing documentation \cite{dinari2023benefits}. 

In this work we contribute to this problem with a case study from home care nursing in Switzerland, dealing with domain-specific vocabulary, local dialects and foreign accents in German. 

In Switzerland, one of the four official languages is German, however, on a daily basis, different Swiss dialects are used. Depending on the specific region, there can be major variations from the standard German, in terms of pronunciation, syntax and vocabulary. These dialects are also used in the professional contexts, and thus, in the context of nursing. Transcription systems, in order to be utilized in practice, do not only need to be able to properly transcribe standard German, but also be able to translate spoken language in different dialects to written standard German. One particular case of nursing activities is the case of \emph{home care nursing (HCN)}. Nurses visiting their patients at home do not have the clinical infrastructure readily available for the documentation, and thus have an additional need for efficient tools supporting their workflows. 

An additional challenge encountered in healthcare, next to a global shortage of health professions\footnote{see e.g., https://www.who.int/news-room/fact-sheets/detail/nursing-and-midwifery}, is the data confidentiality. Health data being subject to strong data protection laws, can often only be processed locally, without making access to third-party products via the Internet, or API models. This requires a trade-off between performance of the models, and available computing resources. 

In this case study, we have investigated the following research questions:
\begin{itemize}
    \setlength\itemsep{0em}
    \item Which state-of-the-art transcription models are the most suitable for the context of HCN documentation in Switzerland?
    \item How do the selected models perform in terms of transcribing (a) domain-specific wording, (b) different Swiss dialects, (c) German with a foreign accent?
\end{itemize}

To conduct our experiments, synthetic texts were created in an interdisciplinary team, representing typical sample sentences of HCN documentation, including particular challenges for the system as specific vocabulary.

\section{Methods}
\subsection{Model Selection Process}
In a first step, an online search of different transcription models and products was conducted. The search was guided by the following questions:
\begin{enumerate}
    \setlength\itemsep{-0.2em}
    \item Does the model/software support German?
    \item Can the model/software be executed locally, or do the audio traces need to be transferred to a third-party?
\end{enumerate}

Whereas the initial search resulted in a long list of potential models and products, the vast majority did not support German and/or was based on third-party APIs where the speech data needs to be transferred to. Open source projects that were inactive for a long time were also excluded. 

Finally, two systems underwent an informal test: (a) OpenAI Whisper\footnote{https://github.com/openai/whisper?tab=readme-ov-file} \cite{radford2022robustspeechrecognitionlargescale} and (b) Vosk\footnote{https://alphacephei.com/vosk/}. A first informal test with both systems did reveal major differences in terms of correct transcription of German, and thus the rest of the experiments was conducted only with the OpenAI Whisper model. 

\subsection{Experimental Setup}
Two standardized texts were presented to three different people, who spoke them into a simple Python web application created with Streamlit, and including the OpenAI Whisper model in the backend. The setup was run locally on a GPU Laptop (Lenovo Legion Pro 7 RTX 4090). A screenshot of this web interface is shown in Appendix \ref{sec:appendix2}. 

Different speech patterns were taken into account, including standard German with and without accent, and different Swiss German dialects. The transcriptions were then compared to the original text. One of the samples was intended to describe statements from the client's point of view, the other is structured in such a way that it could be originated from a community nurse. 

\subsection{Speaker Profiles}
The sample transcriptions were made by three different people and covering the following scenarios: 
\begin{itemize}
\setlength\itemsep{-0.2em}
    \item Hd: Standard German without particular accent 
    \item Hda: Standard German with accent (Russian)
    \item Swiss\_dil1: Swiss German dialect 1 (Basel region)
    \item Swiss\_dil2: Swiss German dialect 2 (Bern region)
\end{itemize}

There is one version of each of the two sample texts from Hda, Swiss\_dil1 and Swiss\_dil2, and two versions each from Hd, which originate from the speakers of Swiss\_dil1 and Swiss\_dil2.

The influence of external noise sources (e.g. road noise) was discussed in advance. For data protection reasons, however, sensitive information should never be disclosed in public spaces. In practice, documentation is therefore either carried out in the client's home or in the service vehicle, or possibly also in the HCN organization's offices. Although sources of noise are also conceivable in a domestic setting (e.g., construction noise in a neighboring apartment), these are not the rule and can be avoided by documenting in a different location. Therefore, no specific recording scenario was developed for this in our experimental setup.

\subsection{Sample Sentences}
Table \ref{tab:texts} gives an overview of the sample sentences in the original German text, and a translation to English. The German texts are written in standard German and the Swiss German dialect speakers were asked to translate them to their corresponding dialect while reading them. Even though in standard German, particular words were included, as for example \emph{Urseli}, which is a local term referring to \emph{hordeolum sty}, which in more standard German would be \emph{Gerstenkorn}. 

\begin{table*}[t]
    \centering
   \begin{tabular}{|p{2cm}|p{6cm}|p{6cm}|}
        \hline
        \textbf{Perspective} & \textbf{German} & \textbf{English} \\ \hline
        Client & Ich habe seit 5 Tagen ein Urseli auf dem linken Auge und Schnupfen, deshalb habe ich etwas den Moralischen. Ausserdem macht der Bauch weh, seit ich gestern Poulet gegessen habe. Und meine Rückenprobleme werden auch nicht weniger. & I've had a sty in my left eye and a cold since yesterday, so my morale is a bit low. My stomach has also been aching since I ate chicken yesterday. And my back problems aren't getting any less. \\ \hline
        Community Nurse & Spitexeinsätze neu 2x täglich zur Mobilisation und Pneumonieprophylaxe. Ebenfalls sollen Interventionen zum Verhindern eines Dekubitus miteingebaut werden. Das Richten des Wochendosetts verbleibt vorerst bei der Klientin. Der Ulcus cruris bedarf weiterhin eines regelmässigen Verbandswechsels, Begutachtung durch Wundexpertin geplant. & Home care nursing interventions now twice daily for mobilization and pneumonia prophylaxis; interventions to prevent pressure sores should also be incorporated. For the time being, the client remains responsible for straightening the weekly bed. The leg ulcer still requires regular bandage changes, assessment by wound expert planned. \\ \hline
    \end{tabular}
    \caption{The example sentences for the client perspective as well as the community nurse perspective.}
    \label{tab:texts}
\end{table*}

\section{Results}
The validation of the results was done by manual assessment of one of the authors, who is a HCN professional. Several metrics, such as Word Error Rate (WER), Character Error Rate (CER), and Accuracy, are commonly used to assess transcription performance; however, the most relevant evaluation depends on the specific needs of professionals, particularly their ability to understand the transcribed content, which is challenging to capture in formal metrics. Therefore, and given the limited number of samples in our experiments, a manual expert assessment was conducted. The following subsections provide a qualitative analysis for each recording. The resulting transcripts are listed in Appendix \ref{sec:appendix}, with remarkably different and potentially not understandable words highlighted. These were assessed by the HCN professional, and the results are shown in Table \ref{tab:results}. Spelling mistakes, as for example \emph{Spitexeinsätze} transcribed to \emph{Spitex einsetze} were ignored, as they do not limit the understanding of the sentence.

\begin{table*}[t]
    \centering
   \begin{tabular}{|p{6cm}|p{3cm}|p{3cm}|p{3cm}|}
        \hline
        \textbf{Perspective/Speaker} & \textbf{General Impression} & \textbf{Nr. Remarkably Different Words} & \textbf{Nr. Not Understandable Words} \\ \hline
        Hd: Client & Good & 1 & 0 \\ \hline
        Hd: Community Nurse & Good & 3 & 0 \\ \hline
        Hda: Client & Sufficient & 2 & 1 \\ \hline
        Hda: Community Nurse & Good & 4 & 0 \\ \hline
        Swiss\_dil1: Client & Sufficient & 3 & 1\\ \hline
        Swiss\_dil1: Community Nurse & Not Sufficient & 5 resp. 4 & 5 resp. 3 \\ \hline
        Swiss\_dil2: Client & Good & 2 & 0 \\ \hline
        Swiss\_dil2: Community Nurse & Sufficient & 3 & 1 \\ \hline
    \end{tabular}
    \caption{Overview of the results from the qualitative assessment of the different transcripts.}
    \label{tab:results}
\end{table*}

\subsection{Standard German Without Particular Accent (Hd)} 

\paragraph{Client Sentences:} The sentences were represented correctly with the exception of the word \emph{Schnupfen}, which was transformed to \emph{schnuffen}. It could, however, be understood from the context.

\paragraph{Community Nurse Sentences:} The transcription contained some minor orthographic inaccuracies. For example, “Verwandwechsel” instead of “Verbandswechsel”, which, however, does not affect the contextual understanding.

\subsection{Standard German with Russian Accent (Hda)} 

\paragraph{Client Sentences:} Even though the accent was perceived as relatively strong, the transcription was almost fully correct. Only the expression \emph{den Moralischen} was transcribed as \emph{dämmerlische}, what would unlikely be understandable, as it is a different and unrelated word.

\paragraph{Community Nurse Sentences:} Generally, the sentences were correct, with \emph{Spitex} (Swiss naming for HCN) being written as \emph{Spic-Tex}, \emph{Wochendoses} instead of \emph{Wochendosett} and \emph{Pneumoniefrophylaxe} instead of \emph{Pneumonieprophylaxe}, but the terms remained understandable. Some technical terms were better transcribed than for the Standard German without accent. We assume the clearer pronunciation of the speaker with accent could have impacted this. 

\subsection{Swiss German Dialect 1 (Basel Region) (Swiss\_dil1)}

\paragraph{Client Sentences:} The transcription was generally correct, containing only some imprecise terms, e.g., \emph{Ursel} instead of \emph{Urseli} which remained understandable; and \emph{Bauch weh, seit ich gestern Poulet gegessen habe} transcribed as \emph{Bauch, wie ich gestern Poulet gegessen habe}, which was classified as being more difficult to understand.

\paragraph{Community Nurse Sentences:}
Two recordings were made for this case, however, the overall quality of the results were not sufficient. Some expressions were not understandable anymore, e.g., \emph{Spitexeinsätze neu 2x täglich zur Mobilisation und Pneumonieprophylaxe} turned into \emph{Spitex einsetzt zweimal täglich die Mobilisation und genommen eine Einprophylaxe}. Some terms are not recognisable anymore, e.g., \emph{Ulcus cruris} becomes \emph{Olkoschlorismus}. It is also noticed that the two transcriptions have different results, even though the dialect, speaker, and the text were the same\footnote{A second recording was made due to some indistinctly pronounced words in the first recording.}.  

\subsection{Swiss German Dialect 2 (Bern Region) (Swiss\_dil2)}

\paragraph{Client Sentences:}
The transcription was generally perceived as good. The words that were different were understandable from the context, e.g., for \emph{Schnupfen}, the word \emph{schnodere} was transcribed, which is a dialect word that can be understood in this context. Similarly, \emph{Ursel} can be contextually understood as \emph{Urseli}.

\paragraph{Community Nurse Sentences:}
The sample was much closer to the original text than the other dialect version (Swiss\_dil1). The specific terms were correct or understandable, with the exception of \emph{Ulkus cruris} being transcribed to \emph{Ulkohuskruiris}.

\section{Discussion and Conclusion}
Our results indicate that with a state-of-the-art transcription model, sufficient or good results were obtained for most of the cases. This included a speaker with a foreign accent and two Swiss dialect speakers. At the same time, we noticed that, even though the sample is small, there were differences with transcripts using the same language but different speakers. There is thus a potential speaker variation that needs to be further investigated in future work. We also noticed that the unclear pronunciation of the words by some speakers might have potentially impacted the results. Therefore, we plan to further experiment with that in the future. In the current experiment, only two of the Swiss dialects (i.e., from the Bern and Basel region) were considered. We want to extend our research to more different dialects, including in particular also the dialect from the Valais region. Furthermore, it should be clarified whether sociolects or demographically driven language variations (such as youth language or varieties primarily found in the older generation) also have an impact. Finally, we have worked in the presented experiments with an out-of-the-box model, and plan to investigate in future work how these models can be customized and improved for the specific use in HCN in the context of Swiss languages.

\section{Limitations}
Different limitations might have impacted the results of our study. On one side, the number of samples and speakers was small and our results will thus need to be confirmed in larger experiments in future work. Given that the original texts were written in standard German, there is a risk that the dialect speakers (even though informed about this risk) did use a wording closer to the standard language than they might have used when speaking naturally. Our evaluation also considered primarily the words themselves. In some cases, the grammar was not correct, which might further impact the understanding of the sentence.

\section{Ethical Considerations}
No original patient data was used, the samples used for the transcription were manually generated and are fully synthetic based on the experiences of the community nurse (student master of science in nursing) involved in the project. The different audio traces were spoken by collaborators of the research project with different backgrounds. All data processing was done locally on hardware at the Bern University of Applied Sciences.

\bibliography{custom}

\begin{thebibliography}{6}
\providecommand{\natexlab}[1]{#1}

\bibitem[{Dinari et~al.(2023)Dinari, Bahaadinbeigy, Bassiri, Mashouf, Bastaminejad, and Moulaei}]{dinari2023benefits}
Fatemeh Dinari, Kambiz Bahaadinbeigy, Somayyeh Bassiri, Esmat Mashouf, Saiyad Bastaminejad, and Khadijeh Moulaei. 2023.
\newblock Benefits, barriers, and facilitators of using speech recognition technology in nursing documentation and reporting: A cross-sectional study.
\newblock \emph{Health science reports}, 6(6):e1330.

\bibitem[{Falcetta et~al.(2023)Falcetta, De~Almeida, Lemos, Goldim, and Da~Costa}]{falcetta2023automatic}
Frederico~Soares Falcetta, Fernando~Kude De~Almeida, Jana{\'\i}na Concei{\c{c}}{\~a}o~Sutil Lemos, Jos{\'e}~Roberto Goldim, and Cristiano~Andr{\'e} Da~Costa. 2023.
\newblock Automatic documentation of professional health interactions: A systematic review.
\newblock \emph{Artificial Intelligence in Medicine}, 137:102487.

\bibitem[{Joshi et~al.(2020)Joshi, Santy, Budhiraja, Bali, and Choudhury}]{joshi-etal-2020-state}
Pratik Joshi, Sebastin Santy, Amar Budhiraja, Kalika Bali, and Monojit Choudhury. 2020.
\newblock \href {https://doi.org/10.18653/v1/2020.acl-main.560} {The state and fate of linguistic diversity and inclusion in the {NLP} world}.
\newblock In \emph{Proceedings of the 58th Annual Meeting of the Association for Computational Linguistics}, pages 6282--6293, Online. Association for Computational Linguistics.

\bibitem[{Kocabiyikoglu et~al.(2022)Kocabiyikoglu, Portet, Gibert, Blanchon, Babouchkine, and Gavazzi}]{kocabiyikoglu-etal-2022-spoken}
Ali~Can Kocabiyikoglu, Fran{\c{c}}ois Portet, Prudence Gibert, Herv{\'e} Blanchon, Jean-Marc Babouchkine, and Ga{\"e}tan Gavazzi. 2022.
\newblock \href {https://aclanthology.org/2022.lrec-1.109} {A spoken drug prescription dataset in {F}rench for spoken language understanding}.
\newblock In \emph{Proceedings of the Thirteenth Language Resources and Evaluation Conference}, pages 1023--1031, Marseille, France. European Language Resources Association.

\bibitem[{Kumar(2024)}]{kumar2024comprehensive}
Yogesh Kumar. 2024.
\newblock A comprehensive analysis of speech recognition systems in healthcare: Current research challenges and future prospects.
\newblock \emph{SN Computer Science}, 5(1):137.

\bibitem[{Radford et~al.(2022)Radford, Kim, Xu, Brockman, McLeavey, and Sutskever}]{radford2022robustspeechrecognitionlargescale}
Alec Radford, Jong~Wook Kim, Tao Xu, Greg Brockman, Christine McLeavey, and Ilya Sutskever. 2022.
\newblock \href {https://arxiv.org/abs/2212.04356} {Robust speech recognition via large-scale weak supervision}.
\newblock \emph{Preprint}, arXiv:2212.04356.

\end{thebibliography}

\appendix

\section{Experiment Transcripts}
\label{sec:appendix}

This appendix contains the transcripts that were generated by the differnet experiments using the OpenAI Whisper Model. Words that are remarkably different (smaller orthographic errors were ignored) are highlighted, and were analyzed by the the HCN expert to generate the results in Table \ref{tab:results}.

\subsection{Standard German Without Particular Accent (Hd)} 

\paragraph{Client Sentences:}
 Ich habe seit fünf Tagen ein Urseli auf dem linken Auge und \hl{schnuffen}. Deshalb habe ich etwas den moralischen. Außerdem macht der Bauch weh, seit ich gestern Poulet gegessen habe. Und meine Rückenprobleme werden auch nicht weniger.

\paragraph{Community Nurse Sentences:}
 Spitex einsetze neu zweimal täglich zur Mobilisation und Pneumonieprophylaxe. Ebenfalls sollen Interventionen zum Verhindern eines Dekubitus mit eingebaut werden. Das Richten des \hl{Wochenlossettes} verbleibt vorerst bei der Klientin. Der \hl{Ulcus Curis} bedarf weiterhin eines regelmäßigen \hl{Verwandwechsels}, Begutachtung durch Wundexpertin geplant.

\subsection{Standard German with Russian Accent (Hda)} 

\paragraph{Client Sentences:}
 Ich habe seit fünf Tagen ein \hl{Ursuli} auf dem linken Auge und Schnupfen. Deshalb habe ich etwas \hl{dämmerlische}. Außerdem macht der Bauch weh, seit ich gestern Poulet gegessen habe. Und meine Rückenprobleme werden auch nicht weniger.

\paragraph{Community Nurse Sentences:}
 \hl{Spic-Tex-Einsätze}, neu zweimal täglich zur Mobilisation und \hl{Pneumoniefrophylaxe}. Ebenfalls sollen Interventionen zum Verhindern eines Decubitus mit eingebaut werden. Das Richten des \hl{Wochendoses} \hl{verbleitet} vorerst bei der Klientin. Der Ulcus Cruris bedarf weiterhin eines regelmäßigen Verbandwechsels, Begutachtung durch Wundexperten geplant.

\subsection{Swiss German Dialect 1 (Basel Region) (Swiss\_dil1)}

\paragraph{Client Sentences:}
Ich habe seit fünf Tagen \hl{Ursel} auf dem linken Auge und Schnupfen, deshalb habe ich etwas \hl{demoralische}. Ausserdem macht mir der Bauch, \hl{wie} ich gestern Poulet gegessen habe und meine Rückenprobleme werden auch nicht weniger.

\paragraph{Community Nurse Sentences v1:}
 Spitex einsetzt zweimal täglich zur Mobilisation auf einem \hl{Neuprofilaxen}. Ebenfalls sollen Interventionen zum Verhinderung einer Dekupitus mit einbauen werden. Das \hl{Richter} vom \hl{Wochenende-Set} verbleibt vorerst bei der Klientin. Der \hl{Olcus Churris} muss weiterhin regelmässig \hl{auf die Amt gewechselt werden}. Begutachtung durch eine Wundexpertin ist geplant.

 \paragraph{Community Nurse Sentences v2:}
 Spitex einsetzt zweimal täglich die Mobilisation und genommen eine \hl{Einprophylaxe}. Es sollen auch Interventionen zum Verhinderung einer Decoupitus mit eingebaut werden. Das Richten des \hl{Wochen-Docs} verbleibt vorerst bei der Klientin. Der \hl{Olkoschlorismus} muss weiterhin regelmässig \hl{durch den Verband} gewechselt werden. Begutachtung durch die Wundexpartin ist geplant.

\subsection{Swiss German Dialect 2 (Bern Region) (Swiss\_dil2)}

\paragraph{Client Sentences:}
 Ich habe seit fünf Tagen ein \hl{Ursel} auf dem linken Auge und \hl{schnodere}. Deswegen habe ich den Moralisch. Ausserdem tut mir der Bauch weh, seit ich gestern Poulet gegessen habe. Meine Rückenprobleme werden auch nicht besser.

\paragraph{Community Nurse Sentences:}
 Spitex einsetzt \hl{neun,} zweimal täglich zur Mobilisation und Pneumonieprophylaxe. Ausserdem sollen Interventionen zum Verhindern eines Dekubitus mit eingebaut werden. Das Richten des Wochendossettes verbleibt \hl{das nächste Mal} bei der Klientin. Der \hl{Ulkohuskruiris} muss weiterhin regelmässige Verbandswechsel haben. Eine Begutachtung durch Wundexpertin ist geplant.

\section{Screenshot from the Web Interface}
\label{sec:appendix2}

\begin{figure}[htbp]
    \centering
    \includegraphics[width=\linewidth]{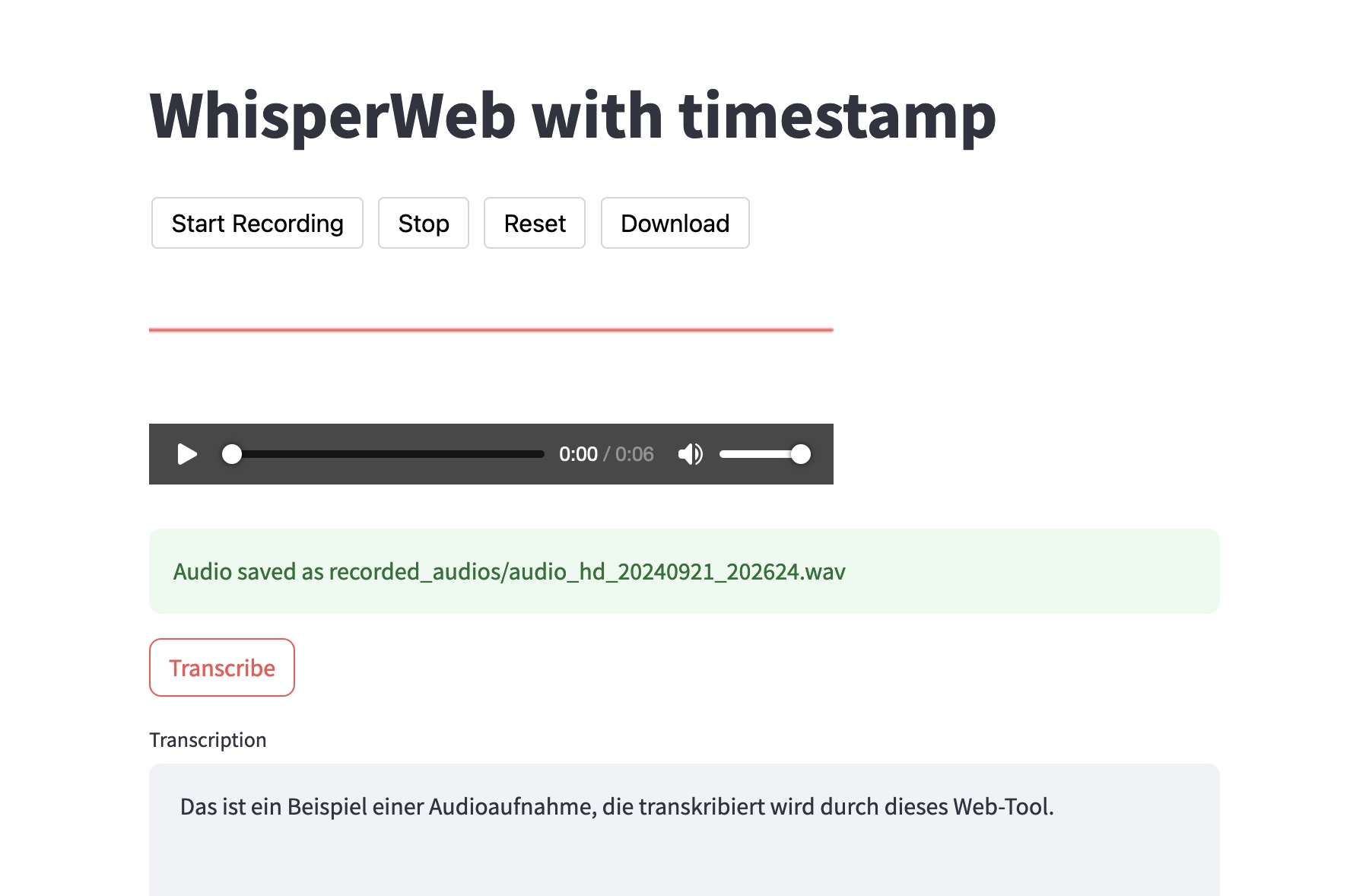}
    \caption{Simple web interface of the Streamlit web application used for recording the transcripts.}
    \label{fig:printscreen}
\end{figure}

Figure \ref{fig:printscreen} shows a screenshot of the web interface used to conduct the transcription expermiments. 

\end{document}